\documentclass[letterpaper, 10 pt, conference]{ieeeconf}
\IEEEoverridecommandlockouts   
\usepackage{graphics}
\usepackage{multirow}
\usepackage{graphicx}
\usepackage{adjustbox}
\usepackage{balance}
\usepackage{hyperref}
\usepackage{amsmath}
\usepackage{breqn}
\usepackage{siunitx}
\usepackage[ruled,vlined]{algorithm2e}
\usepackage{hhline}
\usepackage{todonotes}

\usepackage{todonotes}

\newcommand{\corrS}[1]{\textcolor{black}{#1}}
\newcommand{\revD}[1]{\textcolor{black}{#1}}
\newcommand{\revF}[1]{\textcolor{black}{#1}}
\newcommand{\revL}[1]{\textcolor{black}{#1}}
\newcommand{\revfinal}[1]{\textcolor{black}{#1}}

\title{\LARGE \bf Path and trajectory planning of a tethered UAV-UGV marsupial robotic system*}

\author{S. Mart\'inez-Rozas$^{1}$, D. Alejo$^{2}$, F. Caballero$^{2}$ and L. Merino$^{3}$
\thanks{*This work has been supported by the grants INSERTION PID2021-127648OB-C31 and RATEC PDC2022-133643-C21, funded by MCIN/AEI/ 10.13039/501100011033 and by the “European Union NextGenerationEU/PRTR”.}
\thanks{$^{1}$S. Mart\'inez-Rozas is with Universidad de Antofagasta, Chile. Email: {\tt\small simon.martinez$@$uantof.cl}}
\thanks{$^{2}$D. Alejo and F. Caballero are with Service Robotics Laboratory, Universidad de Sevilla, Spain. Email: {\tt\small dalejo$@$us.es}, {\tt\small fcaballero$@$us.es}}
\thanks{$^{3}$L. Merino is with Service Robotics Laboratory, Universidad Pablo de Olavide, Seville, Spain. {\tt\small lmercab$@$upo.es}}
}

\begin{document}
\maketitle
\thispagestyle{empty}
\pagestyle{empty}

\begin{abstract}
This letter addresses the problem of trajectory planning in a marsupial robotic system consisting of an unmanned aerial vehicle (UAV) linked to an unmanned ground vehicle (UGV) through a non-taut tether with controllable length. \revF{To the best of our knowledge, this is the first method that addresses  the trajectory planning of a marsupial UGV-UAV with a non-taut tether.} The objective is to determine a synchronized collision-free trajectory for the three marsupial system agents: UAV, UGV, and tether. 
First, we present a path planning solution based on optimal Rapidly-exploring Random Trees (RRT*) with novel sampling and steering techniques to speed-up the computation. This algorithm is able to obtain collision-free paths for the UAV and the UGV, taking into account the 3D environment and the tether. 
 Then, the letter presents a trajectory planner based on non-linear least squares. The optimizer takes into account  aspects not considered in the path planning, like temporal constraints of the motion imposed by limits on the velocities and accelerations of the robots\revF{, or raising the tether's clearance. Simulated and field test results demonstrate that the approach generates obstacle-free,  smooth, and feasible trajectories for the marsupial system.}
\end{abstract}

\section{Introduction}
\label{sec:introduction}

A marsupial multi-robot configuration \cite{Murphy1999MarsupiallikeMR} consists of a system in which one robot carries and can deploy one or several other robots. 
One example is the configuration in which an Unmanned Ground Vehicle (UGV) carries an  Unmanned Aerial Vehicle (UAV), which can take off and possibly land back on the UGV when needed. This configuration can be used to combine the strengths of UGVs and UAVs. This UGV-UAV marsupial configuration was used by the Team CSIRO Data61 \cite{CSIRO2021}, reaching second place in the DARPA 2021 Subterranean Challenge \cite{DARPA2021}. Another notable example is the Mars 2020 rover and helicopter \cite{Grip2018GuidanceAC}. The helicopter augments the capabilities of the rover, exploring large areas faster than the rover, providing reconnaissance on target locations and safe to traverse routes. 

The flight time of small UAVs (few tens of minutes) is one of its main limiting factors. To increase their endurance, tethered UAVs fixed to a base station have been proposed \cite{6961531}, but the use of a tether limits the UAV range to a great extent. On the other hand, UGVs have  longer autonomy, but they may not be able to reach certain areas of interest.


\begin{figure*}[t]
    \centering
    \includegraphics[width=\textwidth]
    {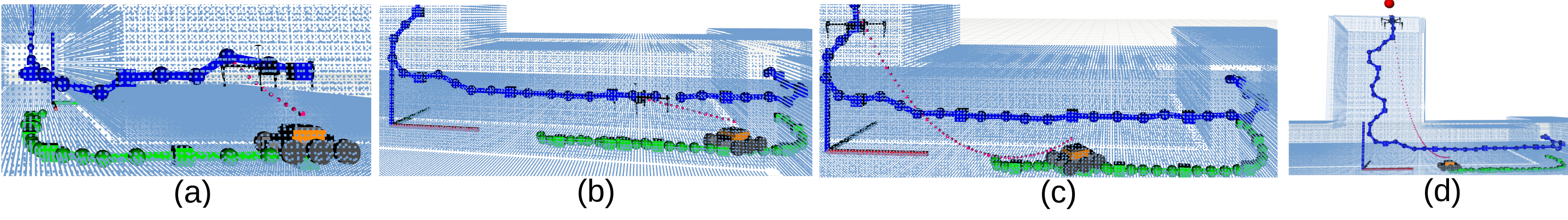}
    \caption{Example of the developed method applied to a Marsupial robotic system. The trajectories of the UAV, UGV are represented in blue and in green, respectively. The tether in a configuration is represented in red. From top to bottom: a) starting configuration. b) configuration after the UAV and UGV turning inside the tunnel; c) the UGV is close to arrive to its final position. d) UAV and UGV in their final positions.}
    \label{fig:intro}
\end{figure*}

\revD{In this letter, we consider a tethered marsupial system, composed by a UAV attached to a UGV with a power cable, see Fig. \ref{fig:intro}. }This configuration enables powering the UAV from the UGV, increasing its endurance and flexibility over systems tethered to a fixed position. The autonomous operation of the system requires considering the UGV-UAV-tether configuration as a whole for planning.

Thus, our goal is the development of motion planning modules for a mobile UGV-UAV-tether system. \revF{To the best of our knowledge, this is the first method that addresses  the trajectory planning of a marsupial UGV-UAV with non-taut tether.} The complexity of this planning problem lies in the dimension of the configuration space, that is, the positions of UAV and UGV, and the tether length. Furthermore, besides considering the collisions of each vehicle, we should also take into consideration the tether. This fact implies a large number of restrictions for the calculation of the trajectory. In our method, we do not assume a taut tether, considering the length of the tie as a control variable. This, together with the position of the UAV and UGV, allows us to be aware of the state of the tie through the mechanical model of the catenary. The main contributions of the letter are:

\begin{itemize}
    \item \revD{A path planning method based on RRT* that generates collision-free paths for UAV and UGV, and lengths of the tether, to achieve a UAV goal configuration.}
    \item \revD{A trajectory planning method based on non-linear optimization that considers smoothness, velocity and acceleration constraints, and optimizes the tether configuration to maximize clearance.} 
\end{itemize}

\revL{We present real experiments of an autonomous tethered UAV-UGV system that show how the provided trajectories are feasible.} The letter is organized as follows. Section \ref{sec:related_work} analyzes the existing works in the literature addressing similar problems. Then, Section \ref{sec:approach} formalizes the problem to solve in this letter. The proposed method for path planning is described in Section \ref{sec:torwardRRT}. This method is used as initial solution for the non-linear optimization problem presented in Section \ref{sec:trajectory-planing} for trajectory planning. The experimental results \corrS{in simulated and real environments} are discussed in \revD{Sections \ref{sec:experiments} and \ref{sec:field-exp}, respectively. Finally, the conclusions and future research directions are detailed  in Section \ref{sec:conclusions}.}

\section{Related Work}
\label{sec:related_work}
Research on tethered systems can be found for Unmanned Underwater Vehicles (UUVs) \cite{LARANJEIRA2020107018}, as the tether is used as a safety recovery device and as a communication link. However, it is usually assumed that the tethers move in free space and thus collisions due to tethers are not considered. The use of tethers has also been applied to UGVs for exploration. In \cite{McGarey2016} a tether is used to anchor the UGV to objects in the environment in order to explore a steep terrain.  In \cite{8794265} a UGV is tethered to a UAV, used as environment sensing assistance and also as an anchor to structures for climbing steep terrain. However, neither of those approaches consider  obstacle avoidance for the tether. 

In contrast, works on motion planning for tethered UAVs and/or tethered UGV-UAV configurations  are scarce. In \cite{6961531}, a power-tethered UAV-UGV team is presented for tasks that require longer operation and increased inspection capabilities. The authors use two independent two-dimensional planners, 
 incorporating the distinct navigation constraints and perception capabilities of each vehicle, the tether constraint, using the RRT* algorithm for path-planning. However, the planning system operates in 2-D planes (the UGV at ground level and the UAV at a given altitude), and it does not consider the length of the tether as a control variable. 

Other approaches that consider aerial robots linked with a tether usually focus on control or estimation aspects. In \cite{9684670}, the authors present a system in which a human is physically connected to an aerial vehicle by means of a cable. A human-state aware controller allows the robot to pull the human toward the desired position including human velocity feedback. In \cite{6907304}, the use of a taut tether is proposed to increase the stability and safety in landing maneuvers. In \cite{schulz_iros_18}, the authors consider also a tether to increase the stability when a UAV flights in confined spaces. In \cite{8848946}, the performance of several flight primitives for a tethered UAV is analyzed. 
The system controls the tether length, but assumes a taut tether to a fixed point. This is a limitation, as it forces the UAV to only access areas with a direct line of sight (LoS) from the fixed point.  \revD{In \cite{drone_marsupial}, the hardware design of a tethered marsupial UGV-UAV platform is described, including the procedures for autonomous landing and a  mission planner for pillar degradation analysis in underground mines. However, the tether is assumed to be taut and the planner generates trajectories with sweeping patterns only for the UAV system.}

A tether state estimation technique is presented in \cite{XiaoSSRR2018}, which describes a relative localization system for a marsupial tethered UGV-UAV pair. The estimation of the tether angles, as measured by the cable’s provider SDK, and a catenary model \cite{BOOKOFCURVES} are used to estimate the relative position of the UAV with respect to the UGV. The same authors propose in \cite{XiaoIROS2018} a PRM planner in which reachable space is constrained by the tether in two ways: avoiding any contacts of the tether with obstacles, or allowing up to two different contact points to extend the effective range. Nevertheless, again the planner assumes a tensed tether at any time, enabling collision checking based on ray-tracing. In \cite{inbookXiao} it is presented a tethered UAV-UGV team that navigates through unstructured or confined spaces. The UAV navigates autonomously and performs as visual assistant for UGV, while UGV is tele-operated. This approach also assumes a taut tether. 

Regarding path planning, \corrS{  \cite{9844242} presents a method for autonomous exploration of unknown cavities in three dimensions (3D) that focuses on minimizing the distance traveled and the length of the unwound tether. The method uses a hierarchical framework composed of a global level of exploration path planning that solves a Traveling Salesman Problem (TSP) and a local level of path planning whose parameters make it possible to adjust the trade-off between the tether unwinding and the path cost. This approach also assumes the tether to be always in a taut condition.}

In \cite{smartinezr2021}, we proposed a trajectory planning method for an UAV tied to a fixed element. The tether length is also used as a decision variable, and a loose tether and its catenary model is used to achieve configurations outside of the LoS of the fixing point. The current work extends this one by proposing solutions to the planning problem for the UGV-UAV-tether system as a whole. This letter presents both, a path and a trajectory planner, to solve such problem. To the best of our knowledge, there are no approaches in the state of the art dealing with path or trajectory planning for a tethered UGV-UAV system considering a controllable loose tether.

\section{Problem Statement}
\label{sec:approach}


We define a state in the state space as the combination of the position of the UGV $\mathbf{p}_g=(x_g,y_g,z_g)^{T}$, the UAV $\mathbf{p}_a=(x_a,y_a,z_a)^{T}$ and the tether length $l$. At any instant, the tether length must be longer than the distance from the UGV to the UAV ($l \geq \|\mathbf{p}_a - \mathbf{p}_g\|$). We obtain 
the shape of the tether using the catenary model \cite{BOOKOFCURVES}, whose parameters can be computed with the Bisection Numerical Method.

The motion planning problem consists in determining the trajectory for the UGV $\mathbf{p}_g(t)$, the UAV $\mathbf{p}_a(t)$ and the tether length $l(t)$ so that the UAV reaches a given goal position, avoiding obstacles and meeting the constraints of the system.

\begin{figure*}[t!]
    \centering
    \includegraphics[width=1.0\linewidth]{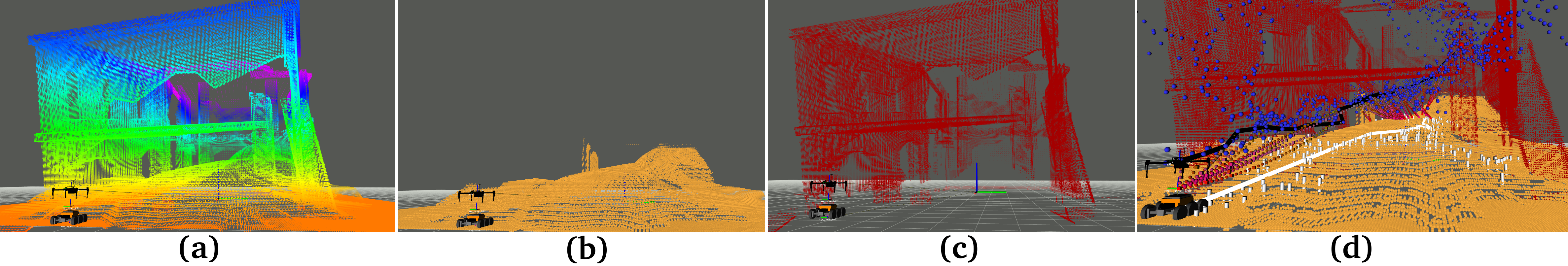}
    \caption{(a) Example of full 3D point cloud (PC) for a scenario used for environment representation. (b) The traversable PC for the UGV computed from full cloud. (c) Obstacle PC for the UGV computed from full PC. (d) In white an example of sampling and computed path over traversable PC.} 
\label{fig:traversable_area}
    \vspace*{-5mm}
\end{figure*}

We discretize the trajectories, so that the states of our problem become the set:
\begin{equation}
\label{eq:traj_params}
    O = \{\mathbf{p}^i_g,\mathbf{p}^i_a,l^i,\Delta t^i\}_{i=1,...,n}
\end{equation}

\noindent where \revfinal{$i$ is a timestep of the trajectory and $n$ is the total number of timesteps.} $\Delta t^{i} = t^{i} - t^{i-1}$ is the time increment between steps $i$ and $i-1$
. \revD{This value is the same for UGV and UAV trajectories. For each $\mathbf{p}^i_a$ , $\mathbf{p}^i_{g}$ and $l^i$, there is a tether configuration $T^i$, given by the catenary model mentioned above. When needed, we discretize it into a set of $m$ positions $\mathbf{p}_{t}=(x_{t}, y_{t}, z_{t})$:}

\begin{equation}
\label{eq:tether}
    T^i = \{\mathbf{p}^j_t\}_{j=1,...,m}
\end{equation}

\section{Path Planning of a Tethered UAV-UGV System Using RRT*}
\label{sec:torwardRRT}

In our first method, we disregard the temporal aspects and, thus, the dynamic constraints of the system, tackling the path planning problem. The objective is to determine the sequence of positions for the UAV and the UGV, and the adequate tether lengths at each step, to attain the UAV goal.

Our approach makes use of the RRT* algorithm \cite{karaman_rrt_star} to solve the problem due to its ability to deal with high-dimensional spaces. However, we do not sample in seven dimensions (UGV, UAV, and tether length) to find the solution because of computation and problem domain reasons. Instead, the tether length \revD{is obtained procedurally in the RRT* \texttt{Steering} algorithm} (see below), reducing the problem dimension to six. Our RRT* cost function is the weighted sum of the total length of the UAV and UGV paths, given that a new node is only accepted if there exists a collision-free catenary connecting the UGV with the UAV.

In order to properly consider the constraints derived from the tethered UAV and UGV configuration, we have \revD{adapted the procedures of RRT* algorithm \cite{karaman_rrt_star}, as described below.}

\subsection{RRT* problem setup}
\label{sec:definition}

The RRT* will plan a path in a six-dimensional space composed by the UGV and UAV position $\mathbf{x}=\{\mathbf{p}_g,\mathbf{p}_a\}$, from an initial state $\mathbf{x}^i$ to a goal UAV state $\mathbf{x}^g$ in which only the aerial robot position is set (the final position of the UGV will depend on the UAV goal and the environment). 
The algorithm makes use of a 3D point cloud to represent the environment in which the robot system plans a path (see Fig. \ref{fig:traversable_area}(a) as an example). This point cloud is processed for efficient obstacle representation for UAV and UGV. First, we use a method similar to \cite{driving_pc} to perform a standard traversability analysis of the input 3D cloud to estimate the points that are traversable by the UGV, given the initial position. This traversable point cloud is used to sample UGV positions (see Fig. \ref{fig:traversable_area}(b) as example). Second, we use a more convenient representation of the UAV environment (the full 3D point cloud) using a 3D grid containing the Euclidean Distance Field (EDF) of the environment \cite{Oleynikova2016SignedDF}. The EDF is queried to get the distance to obstacles from any point
, saving a great deal of computation effort. In our implementation, the EDF is computed offline
, but we can use fast EDF implementations like FIESTA \cite{Han:Iros19} 
for online computation.

\subsection{Sampling process}
\label{sec:sampling}

When obtaining new nodes in free space through the \texttt{Sample} procedure, the UGV position $\mathbf{p}_g$ is sampled from the UGV's traversable points of the point cloud, 
and the UAV position $\mathbf{p}_a$ from the collision-free workspace. This sampling procedure allows us to plan UGV paths not only on flat surfaces but also with elevations \revD{(see Fig. \ref{fig:traversable_area}(d)).}

\subsection{Getting the nearest node}
\label{sec:nearest_node}

To determine the nearest neighbor node in the current tree from a new node, the \texttt{Nearest} procedure considers a weighted sum of the Euclidean distances between the UGV and UAV positions of each node, \corrS{and UGV orientation (UAV orientation is not considered because it is holonomic)}. 

In terms of energy, the maneuvers of the marsupial robot system (once the UAV is flying) can be ordered from higher to lower consumption as follows: 
 UAV translation, UAV hovering,  UGV translation and stopped UGV \corrS{ \cite{8486942}. 
However, since we aim the UAV to reach the goal, our approach prioritizes the movement of the UAV over the UGV (avoiding hovering). Thus, we apply a greater weight factor to the UGV distance. 
The nearest node is the one with the lowest cost.}

\subsection{checkCatenary algorithm}
\label{sec:checkcatenary}

We include this algorithm in the \texttt{Steering} procedure. It checks if there exists a collision-free tether connecting the UGV to the UAV. The algorithm starts with the minimum tether length, i.e. straight line, and increases the length at fixed intervals until a collision-free tether is found or the maximum length is reached. If the collision-free tether is found, the states $\mathbf{p}^i_a$ and $\mathbf{p}^i_g$ are considered as feasible for the length $l^i$, and the algorithm returns \texttt{true}.

\subsection{Steering process}
\label{sec:steering}

The \texttt{Steering} procedure, \corrS{described in Algorithm \ref{alg:rrt_algorithm}}, has been adapted in order to facilitate the RRT* tree growth, as well as minimizing energy consumption of the marsupial system as described in Section \ref{sec:nearest_node}. According to this, three modes have been defined, from highest to lowest priority: 

\begin{enumerate}
     \item The first mode steers the UAV position component only, and the UGV position is fixed. If the UAV is steered without collision and  \texttt{checkCatenary} is \texttt{true}, then the \emph{new node} is saved.
     \item If the first mode cannot provide a viable node, the second mode is executed. The second mode steers UGV and UAV positions in such a way that if their poses are feasible and also \texttt{checkCatenary} is \texttt{true}, the \emph{new node} is saved. Otherwise, the third mode is executed.
     \item The last mode just steers the UGV and considers the UAV fixed. Then, if the UGV pose is feasible and \texttt{checkCatenary} is \texttt{true}, the \emph{new node} is saved. 
\end{enumerate}
 Finally, if a \emph{new node} was not found in any one of three modes, a new \emph{random node} is calculated using \texttt{Sampling}.

\begin{algorithm}[!t]
\small

\label{alg:rrt_algorithm}
\SetKwProg{rrtalgorithm}{steering($X_{nearest}$, $X_{rand}$)}{}{end}

\SetAlgoLined
\rrtalgorithm{}{
$X_{new} \leftarrow steerUAV(X_{nearest},X_{rand})$
    
    \If{obstaclesFree($X_{nearest}$,$X_{new}$) \&\& checkCatenary($X_{new}$)}{
      \Return TRUE, $X_{new}$;
    }
    $X_{new} \leftarrow steerUAVandUGV(X_{nearest},X_{rand})$
    
    \If{obstaclesFree($X_{nearest}$,$X_{new}$) \&\& checkCatenary($X_{new}$)}{
      \Return TRUE, $X_{new}$;
    }
    $X_{new} \leftarrow steerUGV(X_{nearest},X_{rand})$
    
    \If {obstacleFree($X_{nearest}$,$X_{new}$) \&\& checkCatenary($X_{new}$)}{
      \Return TRUE, $X_{new}$;
    }
    \Return FALSE;
}
 \caption{\corrS{ Steering process to get new node $X_{new}$.   Function return TRUE if $\mathbf{X_{new}}$ is feasible. 
 $\mathbf{X_{new}}=\{\mathbf{p}_g,\mathbf{p}_a\}$ save feasible position for UGV and UAV.}}
\end{algorithm}
\subsection{Obstacle Free test}
\label{sec:obstacles_free}
\revF{The \texttt{ObstacleFree} process checks the feasibility (free-collision state) of the new node, and also if the system can be moved from the \emph{nearest node} configuration to the \emph{new node}}. To this end, we interpolate the whole state between those two nodes and check for collisions and the existence of a catenary in each one of the interpolated states.


\section{Trajectory Planning for a Tethered UAV-UGV System Using Non-Linear Optimization}
\label{sec:trajectory-planing}

The result of the path planning method from the previous section is a sequence of collision-free robot positions and tether lengths $\{\mathbf{p}^i_g,\mathbf{p}^i_a,l^i\}_{i=1,...,n}$, \revfinal{where $n$ is the trajectory size}. This sequence does not include the time-related information in (\ref{eq:traj_params}), and neither considers time or safety constraints. In this section, we present a trajectory planning approach that improves that path by solving a non-linear optimization problem \revfinal{with dimension $8n$.}

\subsection{Initial trajectory estimation}
\label{sec:trajectory_estimation}

The first step consists in adding time information to the initial path as the initial trajectory solution. For that, we take into account two main aspects.

The first one aims to adjust the initial sequence to have a set of equidistant waypoints. RRT* creates \corrS{new} nodes at a given distance (epsilon) from their parents, which is convenient for us, as we aim to sample the trajectory at regular intervals. However, because of our proposed \texttt{Steering} process (Section \ref{sec:steering}), we can get packed waypoints at the same position in the case of the UGV. To avoid this, whenever we detect some grouped points we spread them evenly to reach the next non-grouped point. Note that as we check for feasibility in the nodes connecting consecutive states (see Section \ref{sec:obstacles_free}), the new points are feasible. Furthermore, for each point, the tether length is calculated through the \texttt{checkCatenary} procedure.

Then, the $\Delta t^{i}$ values are initialized. To do so, scalar constant speeds, $v_g$ and $v_a$, are imposed over UGV and UAV path respectively. Both trajectories must use the same $\Delta t^{i}$ to make UGV and UAV reach the waypoint at the same time. Thus, $\Delta t^{i}$ value for each state of (\ref{eq:traj_params}) is the largest value between 
 $\|\mathbf{p}^i_g - \mathbf{p}^{i+1}_g\|/v_g$ and $\|\mathbf{p}^i_a - \mathbf{p}^{i+1}_a\|/v_a$.


\subsection{Casting the problem as a sparse non-linear optimization}

Based on the discretization of the states \revD{formulated in } (\ref{eq:traj_params}), our problem consists in determining the values of the variables in $O$ that optimize \revD{the following function $f(O)$}:

\begin{equation}
     O^* = \arg \min_O f(O) = \arg \min_O  \sum_{i,k} \gamma_k  || \delta_k^i(O) ||^2
\end{equation}

\noindent $O^*$ denotes the optimized collision-free trajectory for UAV, UGV, and tether from start to goal configurations. $\gamma_k$ is the weight \corrS{factor for} each component $\delta_k^i(O)$ (known as residual \corrS{ for each $k$ constraint)} of the objective function. Each component takes values between [0-1] and encodes a different constraint or optimization objective of our problem, and will be presented next. \revfinal{
We include the constraints into cost functions as soft constrains, as including them as non-linear hard constraints would increase the complexity of the minimum search
}. The weights values are adjusted empirically and their selected values are shown in Section \ref{sec:synthetic_simulations}. Besides, each component should be evaluated in all the timesteps $i$ of the trajectory 
. These components are local with respect to $i$, as they depend on a few consecutive states in general. Consequently, our optimization problem can be solved with non-linear sparse optimization algorithms. In particular, we use \emph{Ceres-Solver} \cite{ceres-solver} as 
back-end. 


\subsection{Constraints and Objective Function}

We represent the constraints in the problem as penalty costs in the objective function. The proposed constraints are related to the tether length, the limits on velocity and acceleration, and the equidistance among consecutive robot poses. In addition, we aim to maximize the distance from obstacles to each agent, minimize the execution time of the trajectory, and maximize its smoothness. We use the Cauchyloss (robust loss function) as a robust kernel in order to reduce the influence of outliers\revfinal{, which might appear in the map generation for instance,} in the solution \cite{GALLEGO2021174}. 



\subsubsection{Equidistance among consecutive states}
\label{sec:equi_constraints}
The optimization process might unevenly move UGV and/or UAV positions within the planning space in \revD{the search of the minimum cost. }
To avoid it, we force them to keep a given distance among states, $\rho_{eg}$ for the UGV and $\rho_{ea}$ for the UAV. They are obtained by dividing the distance between the initial and goal points along the path by the number of steps. In this section we formulate the UGV case, $\delta^i_{ea}$ is calculated similarly.

\begin{eqnarray}
\label{eq:equidistance}
     \delta^i_{eg} =  \|\boldsymbol p^{i+1}_g - \boldsymbol p^i_g\|\ -\ \rho_{eg}
\end{eqnarray}


\subsubsection{UAV/UGV Obstacle avoidance}
This constraint penalizes the UAV/UGV states whose distance to the nearest obstacle $d^i_{oa}(d^i_{og})$ is closer than a safety distance $\rho_{oa}(\rho_{og})$, which depends on the size of the UAV/UGV.

\begin{eqnarray}
  \label{eq:eq_uav_obst}
  \delta^i_{og} &=& \left \{
      \begin{array}{cc}
      \rho_{og}-d^i_{og} & ,\textrm{if}\  \  d^i_{og} \ < \ \rho_{og}\ \\
      0  & ,\textrm{otherwise}
  \end{array}
  \right .
\end{eqnarray}


\subsubsection{Tether obstacle avoidance}

\revD{This constraint penalizes the proximity to obstacles of the $m$ samples in which the tether is discretized. The tether state is computed by solving the catenary model for $l^{i}$ between $\mathbf{p}^i_g$  and $\mathbf{p}^i_a$, according to (\ref{eq:tether}). We compute the distance to the nearest obstacles of each sample of the tether, $d^i_{ot,j}$, and the residual as the sum of the inverse nearest distances.} We increase the weight of those samples closer than a safety distance $\rho_{ot}$ to guarantee higher costs in these cases using $\rho_{j}= \beta$, with $\beta >> 1$.

\begin{eqnarray}
  \label{eq:eq_teher_obst}
  \delta^i_{ot} &=& \sum_{j=1}^{m} \frac{\rho_{j}}{d^i_{ot,j}}  , \  \rho_{j} = \left \{
      \begin{array}{cc}
      1 & ,\mathrm{if}\  \  d^i_{ot,j}  >  \rho_{ot}\ \\
      \beta & , \textrm{otherwise}
  \end{array}
  \right .
\end{eqnarray}



\subsubsection{UGV traversability}
It is necessary for the UGV to remain in traversable areas. To this end, we impose a $\delta_{trav}$ penalty to the UGV states that are away a distance \mbox{$d^i_{trav}>\rho_{trav}$} from the traversable area. Thus, $d^i_{trav}$ is the distance to the closest point into the traversable point cloud. 
 \begin{eqnarray}
   \label{eq:eq_trav}
   \delta^i_{trav} =\left \{
   \begin{array}{cc}
       d^i_{trav} \ - \ \rho_{trav} & ,\textrm{if}\  \  d^i_{trav} \ > \ \rho_{trav}\ \\
      0  & , \textrm{otherwise}
       \end{array}
      \right .
\end{eqnarray}


\subsubsection{Smoothness}

This constraint is in charge of avoiding abrupt direction changes for UGV and UAV trajectories.
From three consecutive positions, two vectors are calculated. The first between $i$ and $i-1$ states, and the second between $i+1$ and $i$ states. Then, we calculate the angle between these vectors, $\theta_g$ for the UGV, and  $\theta_a$ for the UAV. The residuals $\delta_{s g}$ and $\delta_{s a}$ penalize the states whenever the angle between vectors exceeds a given threshold $\rho_{s g}$ and $\rho_{s a}$ respectively. 

\begin{eqnarray}
   \cos\theta_g  &=& \frac{(\mathbf{p}^i_g-\mathbf{p}^{i-1}_g) \cdot (\mathbf{p}^{i+1}_g-\mathbf{p}^i_g)}{\|\mathbf{p}^i_g-\mathbf{p}^{i-1}_g\|\|\mathbf{p}^{i+1}_g-\mathbf{p}^i_g\|}
   \\
%
  \delta^i_{sg} &=&\left \{
      \begin{array}{c}
      1 - \cos(\theta_g),\ \textrm{if}\   |\theta_g| > \rho_{s g}\ \\
      1 - \cos(\rho_{s g}),\ \textrm{otherwise}
      \end{array}
      \right .
\end{eqnarray}

\begin{figure*}[t]
    \centering
    \includegraphics[width=\textwidth]{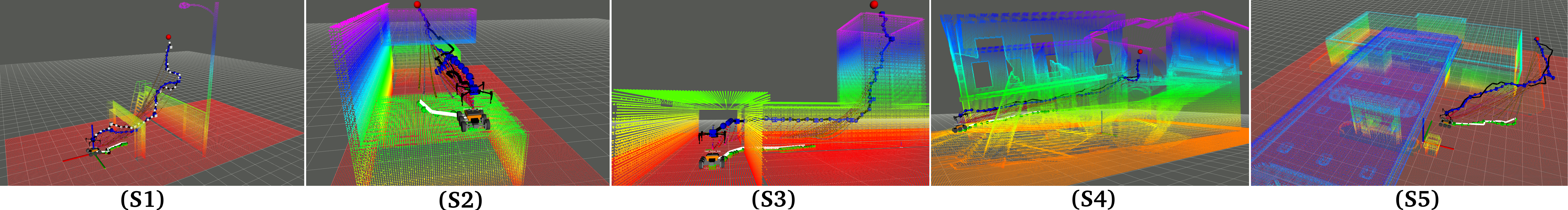}
    \caption{Scenarios considered for validation.S1: Open/constrained space with arc as obstacle. S2: Narrow/constrained space with denied access to UGV. S3: Confined space with outlet duct for UAV. S4: Collapsed Fire Station. S5: Open space gas station.}
    \label{fig:exp_scenarios}
    \vspace*{-5mm}
\end{figure*}





\subsubsection{Velocity}
This constraint relates two consecutive positions and the timestep between them, for the UGV and the UAV. It keeps the speed for both the UGV and the UAV during the optimized trajectory as constant as possible. Then, the error $\delta^i_{vg}$ corresponds to the difference between the computed states velocity and the desired velocity $\rho_{v_g}$. 

\begin{eqnarray}
  \delta^i_{v g} = v^i_{g} - \rho_{v g} 
\end{eqnarray} 

\noindent where $v^i_{g}=\frac{|\mathbf{p}^{i+1}_g-\mathbf{p}^i_g|}{\Delta {t}^{i+1}}$ is 
the average velocity between two consecutive poses for the UGV. 


\subsubsection{Acceleration}
It relates three consecutive positions and their related time steps for the UGV and the UAV, so that the linear acceleration close to zero, minimizing control efforts. 

\begin{eqnarray}
  \delta^i_{ag} = \frac{v^i_g-v^{i-1}_g}{\Delta {t}^{i} + \Delta {t}^{i+1}} 
\end{eqnarray} 

\subsubsection{Unfeasible tether length}
This constraint penalizes unfeasible tether lengths, that is, tethers with length $l^i$ shorter than the Euclidean distance between $\mathbf{p}^i_g$ and $\mathbf{p}^i_a$, $d^i_u$. 

\begin{eqnarray}
  \label{eq:eq_length}
  \delta^i_u =& \left \{
  \begin{array}{cc}
  e^{d^i_u - l^{i} } -1 & ,\textrm{if}\  \  d^i_u \ > \ l^{i} \ \\
  0 &, \textrm{otherwise}
  \end{array}
  \right . 
\end{eqnarray}

\section{Experimental Results in Simulation}
\label{sec:experiments}

Both methods, path and trajectory planners, have been implemented in C++, and the full approach as a planner in Robot Operating System (ROS). The source code is publicly available\footnote{\url{https://github.com/robotics-upo/marsupial_optimizer}}. The \emph{Ceres-Solver} \cite{ceres-solver} has been selected as the back-end to \corrS{solve the optimization problem. We used a 9th gen Intel Core i7 running at 2.20GHz with 32 GB of RAM.}

\label{sec:synthetic_simulations}
To validate the feasibility of the proposed approach, we have conceived a set of experiments in five \revD{synthetic scenarios (S1 to S5), with different degrees of complexity (see Fig. \ref{fig:exp_scenarios}) and also field experiments} . For better understanding, we have included an extended video 
\footnote{\url{https://youtu.be/N-K3yT8Tsxw}}
that includes results on each of the scenarios.  
Our methods can find solutions in environments where approaches that assume a taut tether and/or a fixed UGV pose would fail. As an example, let us consider the output of our method in S3, where the UAV must turn inside a corridor and then enter a chimney to reach the goal, see Fig. \ref{fig:intro}. For that, the UGV should enter the corridor, moving close to the chimney to let the UAV enter.

\revD{For each scenario, we compute a path using the RRT* method in batches of 500 iterations until a solution is found. Then, that solution is used as the initial guess in the optimization step. We set the maximum number of iterations to 50000, obtained by performing experiments in different scenarios, as RRT* usually succeeds before the 1000th iteration.}

We test the methods in two different initial positions for each scenario. The maximum number of optimizer iterations is 1000. All the experiments have been executed 100 times with the same set of parameters detailed below. The weight factors, selected empirically, are in the $[0, 1]$ range.

\begin{itemize}
    \item Weighting factors for UGV: $\gamma_{eg}$ = 0.2, $\gamma_{og}$ = 0.08, $\gamma_{trav}$ = 0.5, $\gamma_{sg}$ = 0.12, $\gamma_{vg}$ = 0.05, $\gamma_{ag}$ = 0.005. 
    \item Weighting factors for UAV: $\gamma_{ea}$ = 0.25, $\gamma_{oa}$ = 0.08, $\gamma_{sa}$ = 0.14, $\gamma_{va}$ = 0.05, $\gamma_{aa}$ = 0.005. 
    \item Weighting factors for Tether: $\gamma_{ot}$ = 0.25, $\gamma_{u}$ = 0.1
    \item Equidistance threshold: $\rho_{eg}$ and $\rho_{ea}$ are computed based on the initial path.   
    \item Collision threshold:  $\rho_{oa}$ = 1.2, $\rho_{ot}$ = 0.1, $\rho_{og}$ = 1.2. 
    \item Traversability threshold: $\rho_{trav}$ = 0.001.
    \item Smoothness threshold: $\rho_{sg} $ = $\frac{\pi}{9}$, $\rho_{sa} $ = $\frac{\pi}{9}$.
    \item Desired velocity (m/s): $\rho_{vg}$ = 1.0, $\rho_{va}$ = 1.0.
    \item Tether avoidance: $\beta$ = 10.0 .
\end{itemize}


The results of the experiments are summarized and detailed in Tables \ref{tab:experiments1} and \ref{tab:experiments2}. 
In order to benchmark the solution, we use as baseline the value of the different metrics prior to optimization, that is, the solution of the RRT* detailed in Section \ref{sec:torwardRRT}. Next paragraphs evaluate the different metrics.

\subsection{Feasibility}
The result of the planners is considered as feasible when the computed path/trajectory is collision free for every agent of the system. This means that $d^i_{og} > \rho_{og}$ and $d^i_{oa} > \rho_{oa}$ for every UGV and UAV state respectively. In case of tether, $d^i_{ot,j} > \rho_{ot}$ for every \revD{sample }in each tether configuration.   

\begin{table*}[t!]
\caption{Results for feasibility, compute time and trajectory time parameters in simulated environments. SI: Scenario;
F: Feasibility; TCI: Time for initial solution [$s$]; TCO: Time for optimized solution [$s$]; VTI: Velocity, initial trajectory [$m/s$]; VTO: Velocity, optimized trajectory [$m/s$]; ATI: Acceleration, initial trajectory [$m/s^2$]; ATO: Acceleration, optimized trajectory [$m/s^2$]} 
\centering
\label{tab:experiments1}
\small
\begin{adjustbox}{max width=\textwidth}
\begin{tabular}{|l|l|ll|llll|llll|llll|llll|}
\hline
\multicolumn{4}{|c}{ALL} &
\multicolumn{8}{|c}{UGV} & 
\multicolumn{8}{|c|}{UAV} \\ 
\hline
\multicolumn{1}{|c}{SI} & \multicolumn{1}{|c}{F} &
\multicolumn{1}{|c}{TCI} & \multicolumn{1}{|c}{TCO} &
\multicolumn{1}{|c}{\begin{tabular}[c]{@{}c@{}}Mean\\ VTI\end{tabular}} & \multicolumn{1}{|c}{\begin{tabular}[c]{@{}c@{}}Max\\ VTI\end{tabular}} &  \multicolumn{1}{|c}{\begin{tabular}[c]{@{}c@{}}Mean\\ VTO\end{tabular}} & \multicolumn{1}{|c}{\begin{tabular}[c]{@{}c@{}}Max\\ VTO\end{tabular}} & \multicolumn{1}{|c}{\begin{tabular}[c]{@{}c@{}}Mean\\ ATI\end{tabular}} & \multicolumn{1}{|c}{\begin{tabular}[c]{@{}c@{}}Max\\ ATI\end{tabular}} &
\multicolumn{1}{|c}{\begin{tabular}[c]{@{}c@{}}Mean\\ ATO\end{tabular}} & \multicolumn{1}{|c}{\begin{tabular}[c]{@{}c@{}}Max\\ ATO\end{tabular}} & 
\multicolumn{1}{|c}{\begin{tabular}[c]{@{}c@{}}Mean\\ VTI\end{tabular}} & \multicolumn{1}{|c}{\begin{tabular}[c]{@{}c@{}}Max\\ VTI\end{tabular}} &  \multicolumn{1}{|c}{\begin{tabular}[c]{@{}c@{}}Mean\\ VTO\end{tabular}} & \multicolumn{1}{|c}{\begin{tabular}[c]{@{}c@{}}Max\\ VTO\end{tabular}} & \multicolumn{1}{|c}{\begin{tabular}[c]{@{}c@{}}Mean\\ ATI\end{tabular}} & \multicolumn{1}{|c}{\begin{tabular}[c]{@{}c@{}}Max\\ ATI\end{tabular}} &
\multicolumn{1}{|c}{\begin{tabular}[c]{@{}c@{}}Mean\\ ATO\end{tabular}} & \multicolumn{1}{|c|}{\begin{tabular}[c]{@{}c@{}}Max\\ ATO\end{tabular}} \\
\hhline{|=|=|=|=|=|=|=|=|=|=|=|=|=|=|=|=|=|=|=|=|}
S1.1 & 86.0 & 5.1  & 465.9 & 0.37 & 0.78 & 0.36 & 1.07 & -0.23 & 0.15 & 0.02 & 0.81  & 0.88 & 1 & 0.97 & 1.19 & -0.74 & 0.25 & -0.002 & -0.09 \\
S1.2 & 80.0 & 15.0 & 529.7 & 0.50 & 0.97 & 0.43 & 1.25 & -0.26 & 0.22 & 0.02 & 0.78  & 0.87 & 1 & 0.91 & 1.19 & -0.72 & 0.32 & -0.002 & -0.08 \\
S2.1 & 94.5 & 0.4  & 123.6 & 0.11 & 0.31 & 0.13 & 0.91 & -0.11 & 0.16 & 0.02 & 0.96  & 0.92 & 1 & 1.01 & 1.15 & -0.71 & 0.06 & 0.002  & 0.01  \\
S2.2 & 95.0 & 0.5  & 109.3 & 0.27 & 0.70 & 0.19 & 1.00 & -0.15 & 0.15 & 0.03 & 1.09  & 0.92 & 1 & 0.99 & 1.09 & -0.70 & 0.11 & 0.000  & 0.07  \\
S3.1 & 84.8 & 0.5  & 213.6 & 0.43 & 0.90 & 0.50 & 1.07 & -0.29 & 0.30 & 0.00 & 0.21  & 0.88 & 1 & 1.02 & 1.21 & -0.73 & 0.20 & 0.003  & 0.16  \\
S3.2 & 84.0 & 1.5  & 214.9 & 0.49 & 0.96 & 0.55 & 1.08 & -0.31 & 0.34 & 0.00 & -0.20 & 0.85 & 1 & 0.98 & 1.16 & -0.69 & 0.26 & 0.004  & 0.18  \\
S4.1 & 82.5 & 1.2  & 539.0 & 0.57 & 0.98 & 0.45 & 1.14 & -0.33 & 0.24 & 0.02 & 1.00  & 0.88 & 1 & 0.94 & 1.12 & -0.71 & 0.30 & -0.002 & -0.06 \\
S4.2 & 76.0 & 12.8 & 708.6 & 0.43 & 0.96 & 0.47 & 1.32 & -0.25 & 0.23 & 0.01 & 0.86  & 0.87 & 1 & 0.92 & 1.22 & -0.70 & 0.40 & -0.002 & -0.32 \\
S5.1 & 98.0 & 27.9 & 277.8 & 0.45 & 0.95 & 0.41 & 1.15 & -0.23 & 0.25 & 0.03 & 1.07  & 0.87 & 1 & 0.97 & 1.27 & -0.64 & 0.25 & -0.004 & -0.14 \\
S5.2 & 95.0 & 26.0 & 447.7 & 0.67 & 1.00 & 0.57 & 1.08 & -0.41 & 0.43 & 0.02 & 1.02  & 0.88 & 1 & 0.92 & 1.11 & -0.68 & 0.39 & -0.004 & -0.12\\
\hline
\end{tabular}
\end{adjustbox}
\vspace*{-2mm}
\end{table*}

An important result is that the RRT* planner is able to find a \revD{ solution} in 100\% of the \revfinal{tested} cases.  On the other hand, the feasibility of the trajectory planner is 86.6\%. \revD{Please note that in 99\% of the unfeasible solutions are due to tether collisions, mainly occurring in cluttered environments with obstacles above and below the tether. The constraints considered by the optimizer might move the solution far from the initial guess provided by  RRT*}. In addition, the tether obstacle avoidance constraint (\ref{eq:eq_teher_obst}) can only be computed \revD{numerically,} making its gradient more sensible to numerical issues. \revD{As a result, the optimizer can produce unfeasible solutions.}

\subsection{Trajectory length, elapsed time and smoothness}

The trajectory planner does not minimize the trajectory length explicitly. The constraints influence the length of the optimized trajectories, possibly increasing their value w.r.t. the initial one. \revD{This is either by moving away from obstacles or by increasing the curvature of the trajectory to smooth it.}

\subsection{Distance to obstacles}
The distance to obstacle statistics shown in Table \ref{tab:experiments2} (DOI, DOO, DCOI, DCOO) for UGV and UAV confirm how the optimized trajectory tends to move away from obstacles when the trajectory itself is closer than the safety distance. For the tether, the DCOO min is on average above the considered safety value $\rho_{ot}$. The DCOI and DCOO mean values are similar for both methods.

\subsection{Velocities and Accelerations}
The results in Table \ref{tab:experiments1} show that the optimized velocities and accelerations (VTO, ATO) are close to the desired values, which are 1.0 $m/s$ and 0 $m/s^2$, respectively. There is a noticeable improvement on these values w.r.t. the initial ones.



\subsection{Computation Time}
Table \ref{tab:experiments1} presents \revD{ the computation time (TCI, TCO) required to compute the initial and optimized trajectories. The results indicate that the RRT* algorithm  obtains a solution in around half a second in confined scenarios, while lasting up to 27.9 seconds in open spaces.} On the other hand, the optimization algorithm requires longer times \revD{ due to the need of repeatedly solving the transcendental catenary equations \cite{BOOKOFCURVES}. However, it is worth noting that CERES solver was configured with a single CPU thread. In this implementation, the computational time can be estimated as the single thread time divided by the number of threads used. That is, we can reduce it an order of magnitude by using ten CPU threads.} 

\begin{table*}[t!]
\caption{Results for trajectory length and distance to obstacles in simulated environments. LIP: Length, initial path [$m$]; LTO: Length, optimized trajectory [$m$]; DOI: Distance to obstacles, initial path [$m$]; DOO: Distance to obstacles, optimized trajectory [$m$]; DCOI: Dist. catenary-obstacle initial solution [$m$]; DCOO: Dist. catenary-obstacles, optimized solution [$m$]}
\centering
\label{tab:experiments2}

\begin{adjustbox}{max width=0.95\textwidth}
\begin{tabular}{|l|ll|llll|ll|llll|llll|}
\hline
\multicolumn{1}{|c}{ALL} &
\multicolumn{6}{|c}{UGV} & 
\multicolumn{6}{|c}{UAV} & 
\multicolumn{4}{|c|}{Tether} \\ 
\hline
\multicolumn{1}{|c}{SI} &
\multicolumn{1}{|c}{LIP} & \multicolumn{1}{|c}{LTO} &  \multicolumn{1}{|c}{\begin{tabular}[c]{@{}c@{}}Mean\\ DOI\end{tabular}} & \multicolumn{1}{|c}{\begin{tabular}[c]{@{}c@{}}Min\\ DOI\end{tabular}} & \multicolumn{1}{|c}{\begin{tabular}[c]{@{}c@{}}Mean\\ DOO\end{tabular}} & \multicolumn{1}{|c}{\begin{tabular}[c]{@{}c@{}}Min\\ DOO\end{tabular}} &
\multicolumn{1}{|c}{LIP} & \multicolumn{1}{|c}{LTO} & 
\multicolumn{1}{|c}{\begin{tabular}[c]{@{}c@{}}Mean\\ DOI\end{tabular}} & \multicolumn{1}{|c}{\begin{tabular}[c]{@{}c@{}}Min\\ DOI\end{tabular}} & \multicolumn{1}{|c}{\begin{tabular}[c]{@{}c@{}}Mean\\ DOO\end{tabular}} & \multicolumn{1}{|c}{\begin{tabular}[c]{@{}c@{}}Min\\ DOO\end{tabular}} & 
\multicolumn{1}{|c}{\begin{tabular}[c]{@{}c@{}}Mean\\ DCOI\end{tabular}} & \multicolumn{1}{|c}{\begin{tabular}[c]{@{}c@{}}Min\\ DCOl\end{tabular}} & \multicolumn{1}{|c}{\begin{tabular}[c]{@{}c@{}}Mean\\ DCOO\end{tabular}} & \multicolumn{1}{|c|}{\begin{tabular}[c]{@{}c@{}}Min\\ DCOO\end{tabular}} \\
\hhline{|=|=|=|=|=|=|=|=|=|=|=|=|=|=|=|=|=|}
S1.1 & 6.3  & 6.3  & 2.26 & 1.45 & 2.28 & 1.55 & 14.6 & 14.2 & 1.83 & 0.65 & 1.87 & 0.80 & 1.15 & 0.31 & 1.14 & 0.25 \\
S1.2 & 10.7 & 10.8 & 3.74 & 1.50 & 3.76 & 1.60 & 18.8 & 18.2 & 2.20 & 0.85 & 2.21 & 0.89 & 1.30 & 0.22 & 1.31 & 0.16 \\
S2.1 & 1.5  & 1.6  & 0.65 & 0.56 & 0.69 & 0.56 & 13.1 & 12.7 & 1.10 & 0.59 & 1.19 & 0.76 & 0.71 & 0.13 & 0.72 & 0.10 \\
S2.2 & 2.7  & 2.7  & 0.75 & 0.58 & 0.75 & 0.58 & 13.4 & 13.1 & 1.11 & 0.56 & 1.16 & 0.67 & 0.75 & 0.17 & 0.75 & 0.13 \\
S3.1 & 8.6  & 8.9  & 0.88 & 0.50 & 0.98 & 0.62 & 17.7 & 17.6 & 0.69 & 0.51 & 0.81 & 0.56 & 0.68 & 0.20 & 0.67 & 0.13 \\
S3.2 & 11.1 & 11.4 & 0.88 & 0.50 & 0.97 & 0.61 & 19.0 & 18.9 & 0.69 & 0.50 & 0.79 & 0.54 & 0.68 & 0.23 & 0.67 & 0.14 \\
S4.1 & 10.8 & 10.8 & 2.12 & 1.39 & 2.12 & 1.40 & 20.9 & 20.7 & 1.09 & 0.53 & 1.16 & 0.65 & 0.80 & 0.22 & 0.74 & 0.14 \\
S4.2 & 15.4 & 16.0 & 1.70 & 1.02 & 1.70 & 1.07 & 27.1 & 26.7 & 1.06 & 0.51 & 1.15 & 0.65 & 0.82 & 0.22 & 0.76 & 0.10 \\
S5.1 & 9.4  & 9.2  & 2.24 & 0.70 & 2.26 & 0.76 & 18.5 & 18.1 & 1.71 & 0.72 & 1.74 & 0.83 & 1.12 & 0.30 & 1.09 & 0.23 \\
S5.2 & 23.7 & 23.3 & 2.52 & 0.59 & 2.53 & 0.63 & 35.2 & 34.9 & 1.65 & 0.58 & 1.66 & 0.65 & 1.17 & 0.25 & 1.14 & 0.22 \\

\hline
\end{tabular}
\end{adjustbox}
\vspace*{-2mm}
\end{table*}

\section{Field Experiments}
\label{sec:field-exp}

\revF{Once the method has been analyzed in different simulation scenarios, we proceed to validate the solution in a real marsupial system using state-of-the-art localization, navigation, and control approaches to execute the proposed trajectories.}

\revD{The marsupial team is commanded to inspect the old theater building of the Pablo de Olavide University, Seville (Spain) (see Fig. \ref{fig:real_trajectory}.(a)). To this end, the UAV should go to an inspection point (IP) above the stage of the theatre, out of the reach of the UGV and not in its LoS. Therefore, the UGV should go to a position that allows the UAV to reach the IP. The experiment is shown in the attached video.}

\subsection{Marsupial System}

\revF{The marsupial system consists of a M210 drone platform from DJI, an ARCO omnidirectional ground robot from IDMind Robotics, and a custom-made automated reel that adjusts the length of the tether. Each subsystem performs its tasks fully autonomously. The robots carry an OS1 LIDAR for localization and navigation, no external motion capture system was used. In contrast, we used our Direct LIDAR Localization method \cite{dll} to estimate their pose.}

\revF{In order to track the trajectory computed by our approach, the marsupial system must perform a coordinated motion of the UGV, UAV, and automated reel. We opt for a loosely-coupled solution based on synchronization. Thus, new waypoints cannot be commanded until all the subsystems confirm reaching the current one. When a subsystem reaches its goal, it waits for the others in case of need.}

\revF{For safety reasons, the maximum commanded speed has been limited to the slowest subsystem, which in this implementation is the automated reel. Thus, the planner has been parameterized as in Section \ref{sec:experiments}, but setting $\rho_{vg}$ and $\rho_{va}$ to $0.25 m/s$ to accommodate the solution to the reel dynamics.}  

\revF{Both robots implement a control scheme based on PIDs and trapezoidal velocity profile for waypoint tracking. The maximum velocity for the trapezoid is set to the planned $\rho_{vg}$ and $\rho_{va}$ respectively, but there are velocity ramps 
that slightly reduce the average speed during the maneuvers.} 



\subsection{Summary of the results}

\begin{figure*}[bt]
    \centering
    \includegraphics[width=\textwidth]{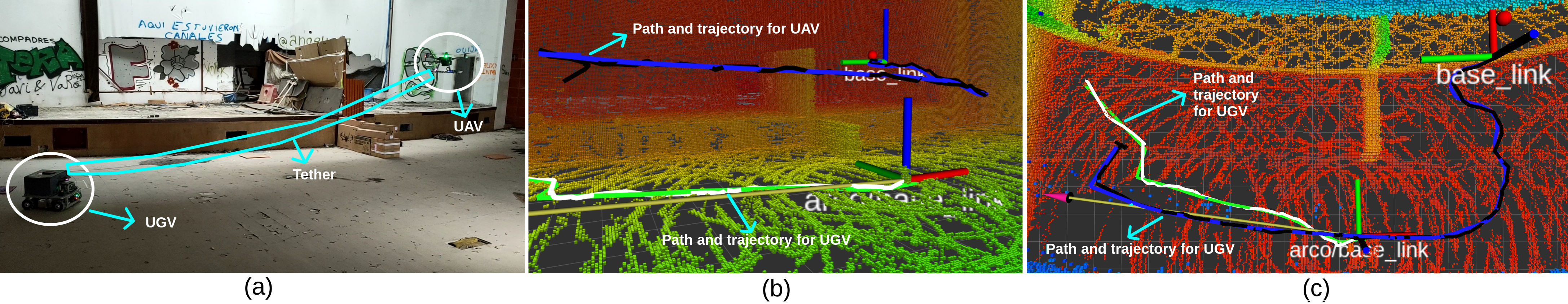}
    \caption{\revD{Validation of the Experiments. (a) Our marsupial system during the experiments at the abandoned theatre. (b) Computed and executed trajectories of the UAV (in blue and black lines, respectively) and UGV (green and white) in the camera view. (c) Same representation in top view.}}
    \label{fig:real_trajectory}
\end{figure*}

\revF{A trajectory for the complete system was computed by the proposed method to guide the system from starting to goal points (see Fig. \ref{fig:real_trajectory}). This trajectory was successfully tracked by the marsupial system in five different experiments at a commanded speed of 0.25 m/s for both platforms, the desired speed ($\rho_v$) in the optimization process. Table \ref{tab:realexperiments2} shows the maximum and average tracking errors in position, velocity and time of each subsystem for one of those experiments. The mean position tracking error of both platforms was below $0.06 m$ in the experiments. The robots' control system implemented a reaching-waypoint-criteria of $0.1 m$ around the objective. Thus, the waypoints located closer than $0.1 m$ were discarded when calculating the errors.}


\revF{The system successfully tracked the trajectory with mean errors below $0.08 m/s$ in velocity and $0.08m/s^2$ in acceleration. The tether tracking errors were also slim during the whole experiment, about $0.01m$ in average. The maximum errors shown in Table \ref{tab:realexperiments2} are mostly produced by the trapezoidal velocity profile scheme for trajectory tracking. The velocity ramps reduce the averaged robot velocity, producing deviations in velocity, acceleration, and time.  In addition, the UGV performs a rotation in place when the angle to reach the waypoint exceeds a given threshold, increasing the execution time. This happened twice in the experiment.}

\revD{All in all, the marsupial system was able to follow the plan without collisions with the tether, as shown in the video.}


\begin{table*}[t!]
\caption{Statistical analysis for trajectory tracking errors (time [$s$], velocity [$m/s$], acceleration [$m/s^2$])}
\centering
\label{tab:realexperiments2}
\begin{adjustbox}{max width=0.8\textwidth}
\begin{tabular}{|c|c|c|c|c|c|c|c|c|c|c|}
\cline{2-11}
\multicolumn{1}{c|}{}              & \multicolumn{4}{c|}{UGV} & \multicolumn{5}{c|}{UAV}             & \multicolumn{1}{c|}{Tether}   \\
\hline              
Measure    & Position XY & Time   & Vel.   &   Acc.  & Position XY & Position Z & Time   & Vel    &   Acc.  & 
Length \\
\hline  
Mean error & 0.03    & 0.46 & 0.08 & 0.08  &
0.05     & 0.06    & 0.35 & 0.05 & 0.05  &
0.01 \\
Max error  & 0.17    & 3.32 & 0.18 & 0.53 &
0.30     & 0.33    & 2.00 & 0.15 & 0.21 &
0.31 \\
\hline
\end{tabular}
\end{adjustbox}
\end{table*}



\section{Conclusions and Future Work}
\label{sec:conclusions}
This paper presented a general path and trajectory planning algorithm for a marsupial system. \revD{Contrary to most existing approaches, our methods do not assume a taut tether, using the catenary model instead. In this way, we can consider both taut and loose configurations.}
\revD{The planning methods have been tested in simulation environments that need of UGV translation and a loose tether to reach the goal position. }

\revD{Despite the complexity} of the problem, our adapted RRT* algorithm provides solutions in few seconds, with a success rate of 100\%. The optimization process generates safer trajectories than planner in terms of distance to obstacles, especially related to UGV and UAV agents. \revD{Finally, we have executed a trajectory generated by the proposed method in field experiments, demonstrating its safety and feasibility. } 

Future work will consider the development of analytical approximations to speed up the computation of the \revD{state} of the catenary model. Analytical approximations of the catenary will improve error gradient estimation. 
 \revfinal{ Finally, some tether collisions planning could be allowed, e.g. with the floor.}


\balance
\bibliographystyle{IEEEtran} 
\bibliography{IEEEabrv}

\end{document}